%% file: main.tex
\pdfoutput=1

\documentclass[11pt]{article}

\usepackage{acl}

\usepackage{times}
\usepackage{latexsym}

\usepackage[T1]{fontenc}

\usepackage[utf8]{inputenc}

\usepackage{microtype}

\usepackage{inconsolata}

\usepackage{graphicx}
\usepackage{subcaption}
\usepackage{soul}

\title{Analyzing Toxicity in Deep Conversations: A Reddit Case Study}

\author{Vigneshwaran Shankaran \\
  GESIS - Leibniz Institute for the Social Sciences \\
  \texttt{vigneshwaran.shankaran@gesis.org} \\\And
  Rajesh Sharma \\
  University of Tartu \\
  \texttt{rajesh.sharma@ut.ee} \\}

\begin{document}
\maketitle

\input{abstract}

\input{introduction}

\input{related_work}

\input{conversation_modeling}

\input{dataset}

\input{rq1}

\input{rq2}

\input{rq3}

\input{rq4}

\input{rq5}

\input{conclusion}

\bibliography{references}

\end{document}

%% file: abstract.tex
\begin{abstract}
Online social media has become increasingly popular in recent years due to its ease of access and ability to connect with others. One of social media's main draws is its anonymity, allowing users to share their thoughts and opinions without fear of judgment or retribution. This anonymity has also made social media prone to harmful content, which requires moderation to ensure responsible and productive use. Several methods using artificial intelligence have been employed to detect harmful content. However, conversation and contextual analysis of hate speech are still understudied. Most promising works only analyze a single text at a time rather than the conversation supporting it. In this work, we employ a tree-based approach to understand how users behave concerning toxicity in public conversation settings. To this end, we collect both the posts and the comment sections of the top 100 posts from 8 Reddit communities that allow profanity, totaling over 1 million responses. We find that toxic comments increase the likelihood of subsequent toxic comments being produced in online conversations. Our analysis also shows that immediate context plays a vital role in shaping a response rather than the original post. We also study the effect of consensual profanity and observe overlapping similarities with non-consensual profanity in terms of user behavior and patterns.
\\ \noindent {\color{red}\textbf{Disclaimer}} This paper contains examples of explicit language that may be disturbing to some readers.
\end{abstract}

%% file: introduction.tex
\section{Introduction}

In recent years, online social media platforms have become ubiquitous in our daily lives. These platforms allow users to connect and communicate with others in real-time. However, the nature of these interactions can vary significantly based on their visibility. Private conversations that take place through online communication channels, such as direct and group messaging, have little effect on society as a whole since they do not influence or challenge the dominant narratives, values, or expectations of society. But conversations occurring in a public setting, such as on social media platforms where users can participate, must adhere to the platform's rules and the general societal norms. However, on the downside, as online social media provide users with a degree of anonymity, this often leads to the emergence of hate speech on the platforms \cite{mondal2017measurement}. Therefore, it is important to moderate content on social media because it helps ensure that the platform remains safe and respectful for all users.
Without moderation, social media can become a breeding ground for spam, hate speech, and other forms of harmful or offensive content. Subsequently, this can create a negative user experience and make the platform less enjoyable for everyone.

\begin{figure*}[t]
\centering
\includegraphics[width=\linewidth]{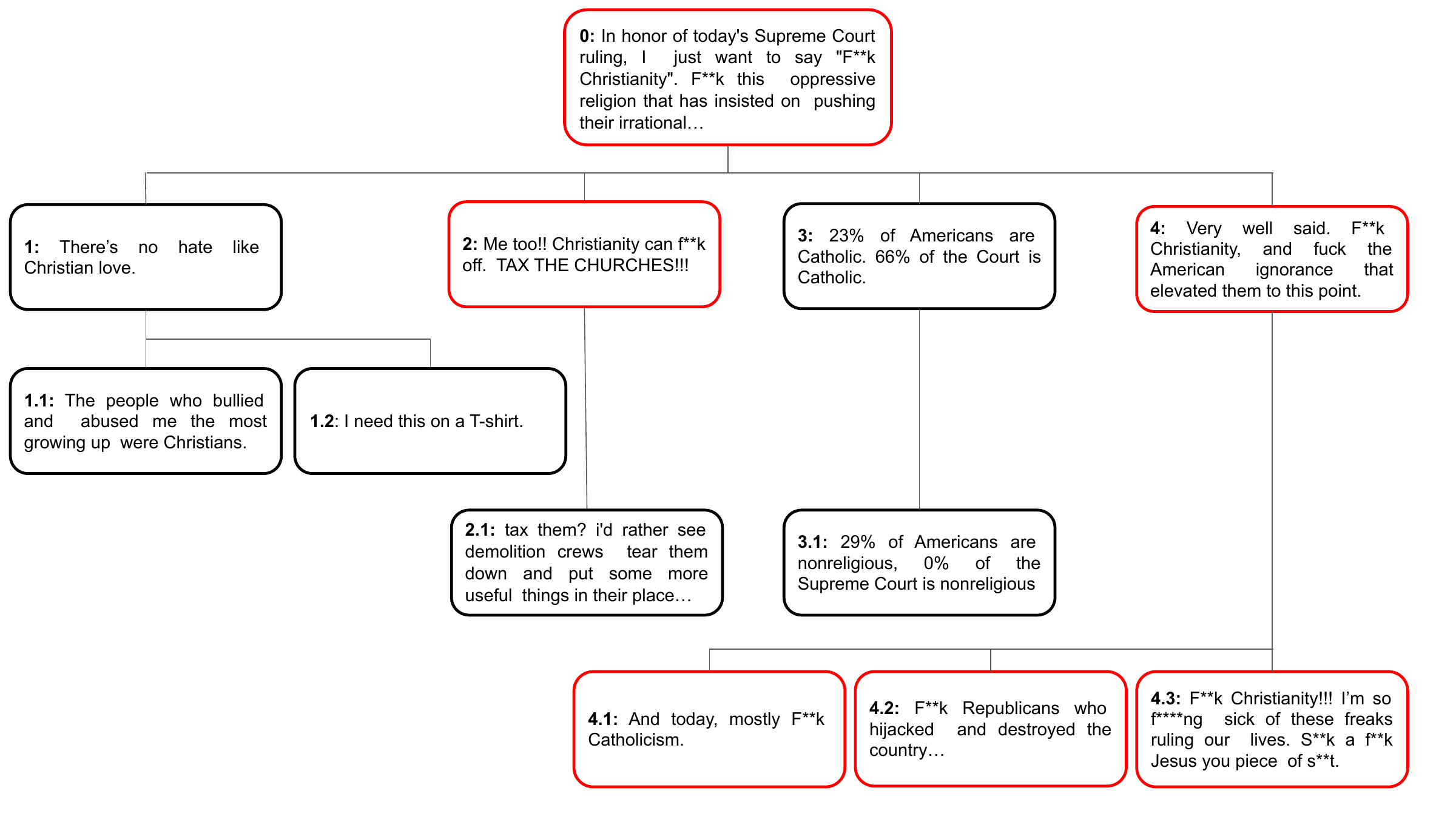}
\caption{Excerpt from a Reddit post. Nodes with red borders signify toxic content as classified by the ToxiGen classifier}
\label{fig1}
\end{figure*}

Given the vast quantity of content, it is difficult to manually examine and identify every instance of hate speech on social media. Consequently, several studies have investigated how to identify hate speech using artificial intelligence. Existing automatic moderation systems use texts sourced from various social media platforms and online encyclopedias manually labeled by human annotators \cite{waseem2016hateful, davidson2017automated, wulczyn2017ex}. The aforementioned studies focus on texts as individual sentences and disregard the notion of conversations that occur in the form of responses to posted content on online social media platforms. Some social media platforms, like Instagram, Youtube, etc., prohibit users from posting nested responses and limit them to a single level. Whereas other platforms like Twitter and Reddit facilitate deeper conversations by permitting nested responses, allowing for discussions similar to real life. Moreover, it is worth noting that certain online communities exhibit distinct norms that may be more permissive or even toxic compared to others. In such cases, new users often find themselves adapting to these established norms \cite{rajadesingan2020quick}. Figure \ref{fig1} is an example of an online conversation sourced from Reddit where the responses to the post are ordered in a hierarchical structure.

This paper focuses on analyzing deep conversations by taking a holistic view of a post and its responses. Specifically, this study centers on the identification of toxic language and patterns within a given post and its subsequent responses. To achieve this, we analyze a dataset consisting of 800 Reddit posts that have garnered over 1 million responses. These posts are sourced from eight distinct subreddits known for hosting lengthy discussions. We use Reddit since a response to a comment or post expresses agreement or disagreement with the original text; it can be viewed as a public conversation. The main contribution of this paper is to find how toxicity disseminates in public conversations on social media platforms that support deep conversations.

Our analysis investigates the following research questions:

\begin{itemize}
\item \textbf{RQ1:} \textit{ Are toxic comments more likely to generate toxic conversations than non-toxic ones?}\\
We find a significant correlation between the toxicity of response and the subsequent toxicity it attracts in terms of responses. See Section \ref{sec:rq1}

\item \textbf{RQ2:} \textit{ Does the toxicity of a response depend on the toxic levels of responses made before it?} \\
According to our analysis, the response immediately preceding a given response has a greater impact on its toxicity compared to the previous responses. See Section \ref{sec:rq2}

\item \textbf{RQ3:} \textit{ How long does a conversation retain its toxic nature?}\\
We observe that toxicity in a conversation generally diminishes within the initial two to three levels of responses and does not endure over time. See Section \ref{sec:rq3}

\item \textbf{RQ4:} \textit{ What effect does toxicity have on user retention in a conversation?}\\
We observe that users exhibit a bimodal pattern in terms of toxicity levels, providing an interesting insight into the dynamics of neutral responses in public conversations. See Section \ref{sec:rq4}

\item \textbf{RQ5:} \textit{Does the toxicity norms of consensual groups differ from non-consensual groups?}\\
We find overlapping similarities between both groups, suggesting identical toxicity norms. See Section \ref{sec:cp}

\end{itemize}

The remainder of the paper is structured as follows. In Section \ref{sec:related_work}, we present relevant works. We define our conversational modeling approach in Section \ref{sec:conversation_modeling}. In Section \ref{sec:dataset}, we discuss the dataset, then in Sections \ref{sec:rq1} through \ref{sec:cp}, we address the research questions and in Section \ref{sec:conclusion}, we provide concluding remarks.

%% file: related_work.tex
\section{Related Work} \label{sec:related_work}

In this section, we discuss existing literature on topics pertinent to our research. We explore two distinct yet interconnected domains relating to our work: toxicity detection and user behavior analysis.

\subsection{Toxicity Detection}

The early efforts to address abusive language, such as \cite{yin2009detection}, employed supervised classification with n-gram, manual regular expression patterns, and contextual characteristics based on prior phrases' abusiveness. Diverse datasets on topics like racism, feminism, and misogyny were gathered from online platforms like Twitter and Reddit \cite{waseem2016you, founta2018large, waseem2016hateful}. Additionally, datasets, including examples of personal attacks and toxic messages from Wikipedia's Talk pages, were used to analyze factors like hate speech, toxicity, and community sentiment on these platforms \cite{wulczyn2017ex}. However, models trained on these datasets struggled with implicit hate, leading \cite{elsherief2021latent} to source tweets from extremist groups for labeled implicit toxicity.

Automatic hate speech detection's inherent complexity poses challenges, as studies highlight difficulties arising from language variations and biases toward certain identity groups in automated systems \cite{sap2019risk, nobata2016abusive, davidson2019racial, xia2020demoting}. Researchers addressed bias concerns, introducing metrics specific to identity performance evaluation \cite{dixon2018measuring, borkan2019nuanced}. Approaches like functional test suites \cite{rottger2020hatecheck}, human-in-the-loop dataset generation \cite{vidgen2020learning}, and multi-task learning \cite{vaidya2020empirical} aimed to enhance model robustness. Additionally, \cite{barikeri-etal-2021-redditbias} used embedding projections to mitigate unintended bias. While previous works focused on single textual entities, \cite{ghosh-etal-2018-sarcasm} analyzed sarcasm using LSTM and attention models, and contextual analysis in fake news detection incorporated hyperbolic geometric representations and Fourier transformations \cite{grover2022public}. An extension of this contextual approach was adapted for implicit hate speech detection \cite{ghosh2023cosyn}.

\subsection{User Behavior Analysis}

While studies have improved toxicity detection models, a different line of work has analyzed social media platforms to unearth user behavior and information dissemination patterns in human communications in cyberspace. \cite{aumayr2011reconstruction} pioneered methods for reconstructing reply behaviors in threads, employing a classification approach that integrates both content and non-content features. \cite{leavitt2014upvoting} delved into the production of news on a hurricane-related subreddit, investigating categories of news likely to receive more upvotes. Notably, followers of influencers disseminating deceptive content on social media have been observed to employ uncivil language and become more effectively polarized  \cite{guldemond2022fueling}.

In addition to broader analytics on social media platforms, there has been a noteworthy surge in literature specifically examining toxicity within these platforms. A study on Twitter user data by \cite{Yousefi2023TowardsDA} highlighted how various content types and contexts can impact the spread of toxicity within online communities. Furthermore, a cross-platform study indicated that Reddit tends to exhibit higher toxicity levels than other platforms, such as Twitter and Parler \cite{Noor2023ComparingTA}, thereby influencing our choice of Reddit as a data source. An insightful examination of Reddit users revealed that usernames serve as useful predictors for author profiling, as users with toxic usernames tend to generate more toxic content than others \cite{urbaniak2022namespotting}.

The works closest to ours involve studies that have analyzed more than one post at a time with respect to toxicity. For example, \cite{bakshy2012role} delved into information diffusion on Facebook, seeking insights into the factors influencing post reach. \cite{aleksandric2022twitter} conducted a longitudinal study on Twitter, uncovering behavioral reactions of toxicity victims, including avoidance, revenge, countermeasures, and negotiation. \cite{kumar2010dynamics} explored the dynamics of online conversations, analyzing patterns in human exchanges on social media, scrutinizing thread depth, and correlating their findings with Heap's law \cite{heaps1978information}. While previous works have investigated user interaction behavior in public conversations, such as continuity \cite{kim2023predicting} and new \& re-entry \cite{wang2022successful}, our approach distinguishes itself by analyzing deep public conversations using a tree-based conversation modeling approach. This methodology enables us to capture the intricate interplay of context and toxicity without imposing limitations on contextual depth, providing valuable insights into how toxicity shapes the buildup of conversations in a public setting.

%% file: conversation_modeling.tex
\section{Conversation Modeling}\label{sec:conversation_modeling}

By and large, in online social media or content-sharing platforms, users post content and receive responses in the form of likes, replies, or comments. In particular, responses in the form of comments facilitate the generation of conversations in a public setting which can be either deep or shallow, depending on the platform. We refer to comments as responses henceforth in this paper. Platforms like Youtube, Instagram, and Facebook allow for shallow conversations with little engagement since only a single level of responses is permitted. However, Twitter and Reddit promote a higher level of engagement owing to their support for deep conversations. Due to the inherent conversational structure of deep conversations, they can be represented as a generic tree model where the post and responses are represented as nodes. Figure \ref{fig1} depicts a deep conversation in which the post or content by itself is the root node, and the remaining nodes are responses. The responses of each individual node are represented as child nodes. Note that every conversation has the possibility of containing toxic responses due to the acceptance of profanity in the chosen subreddits. To proceed further with our analysis, we define the following:

\begin{itemize}

\item \textbf{Opinions :} Opinions can be of any reply with relation to the node's content. In other words, an opinion does not necessarily need to be supportive or against the node's content as it could be neutral, too. Opinion is determined by the number of children nodes of a response. More responses (children) correspond to more opinions for any given response or post. 

\item \textbf{Engagement :} This represents a node's (response) maximum depth. A deeper branch of response indicates that there is a higher level of interest and participation in the discussion, encouraging users to continue contributing their thoughts and opinions on the topic. As a result, a node with a greater depth is often seen as a sign of successful engagement and active participation.

\item \textbf{Toxic Accumulation :} Every node comprising of text is treated as a single entity and a toxicity score is computed based on it. The toxic accumulation score is computed by averaging the toxicity scores of all the nodes that branch from the given node. This indicates the level of toxicity produced by a node. A higher score indicates that the response contributed to the emergence of toxic conversations. Toxic accumulation can be defined by the following equations:

\begin{equation}
\centering\resizebox{.85\columnwidth}{!}{
    $TA_{SubTree}(Node_i) = \sum_{j}^{\#Children} TA({Node_i}_j)$
}
\label{eq1}
\end{equation}

\begin{equation}
\centering\resizebox{.85\columnwidth}{!}{
    $TA(Node_i) = \frac{Toxicity(Node_i) + TA_{SubTree}(Node_i)}{\#Children + 1}$
}
\label{eq2}
\end{equation}

In Equation \ref{eq1}, $TA_{SubTree}(Node_i)$ denotes the contribution of toxicity by subtree formed by $Node_i$. It must be noted that Equations \ref{eq1} and \ref{eq2} are recursive and complement each other. Equation \ref{eq2} computes toxic accumulation of $Node_i$. Therefore, $TA({Node_i}_j)$ in Equation \ref{eq1} refers to the summation of toxic accumulation of all the children $j$ of $Node_i$.

\end{itemize}

These definitions help us better understand user participation in the presence of toxic comments by establishing the relationship between engagement and opinions as well as toxicity and toxic accumulation.

%% file: dataset.tex
\section{Dataset}\label{sec:dataset}

We use Reddit posts for analysis as the platform facilitates deep conversations and provides a large quantity of data with diverse discussions, allowing us to analyze the nature of deep conversations on the platform thoroughly. According to Reddit’s algorithm\footnote{\url{https://github.com/reddit-archive/reddit/blob/master/r2/r2/lib/db/_sorts.pyx}}, top posts are categorized by upvotes and user participation among other factors and therefore are prone to generate deep conversation, which suits our purpose. The top 100 posts as of October 12th, 2022 were chosen from the following eight subreddits: r/atheism, r/confidentlyincorrect, r/Fuckthealtright, r/antifeminists, r/facepalm, r/4chan, r/tifu and r/RoastMe. As the titles of the subreddits imply, one may assume that we chose them for the controversial aspects they might encourage, however, these subreddits were chosen for free discourse, debate, and \textbf{communal acceptance of profanity} in the respective subreddits. We extracted the posts and comments while preserving the hierarchy of the conversation. Each comment’s text was collected along with its metadata, including the score, anonymized author ID, comment ID and time stamp, and stored as individual JSON files\footnote{\url{https://github.com/s-vigneshwaran/Reddit-Case-Study-Data}}.

\begin{figure}[ht]
\centering
\includegraphics[width=\linewidth]{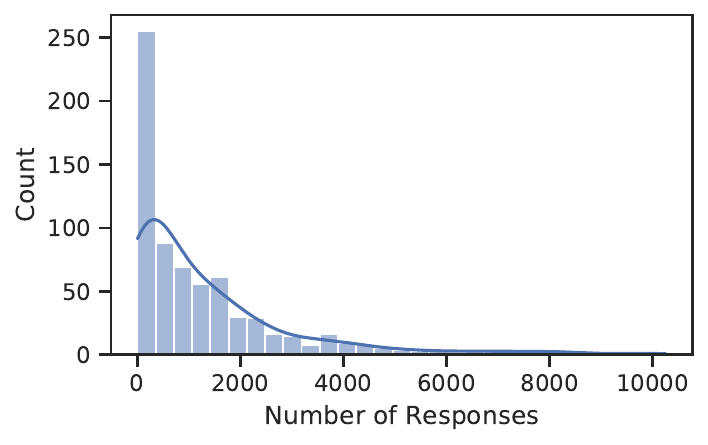}
\caption{Distribution of the number of responses to posts}
\label{fig2}
\end{figure}

Reddit allows multiple forms of content as the body of a post, like HTML, image, audio, gif, etc. A manual examination of 10 randomly selected posts from each subreddit revealed that many posts included pictures with text in some way, be it a quote, a screenshot of a news article, or a Twitter post, etc. We extracted text from the images of such posts and included it in the main body of the post manually because the conversation generated has a direct effect on the post's content. Upon inspection of the entire data \textit{(after extracting texts from images)}, we found that the posts had a median of 52 words, and the responses had a median of 16 words. Figure \ref{fig2} shows the distribution of the number of responses to posts signifying that majority of post attracts fewer than 2000 responses. It must be noted that we exclude posts from r/RoastMe and use only 700 posts for Sections \ref{sec:rq1} to \ref{sec:rq4} since they are targeted towards analyzing non-consensual profanity. The remaining 100 posts from r/RoastMe has been utilized in Section \ref{sec:cp} to study consensual profanity patterns. For a detailed breakup, refer to Table \ref{table1}.

\begin{table}[ht]
\centering
\resizebox{0.85\columnwidth}{!}{
\begin{tabular}{lr}
\hline
\textbf{Subreddits}    & \multicolumn{1}{l}{\textbf{Comments}} \\ \hline
r/atheism              & 116,927                               \\
r/confidentlyincorrect & 189,124                               \\
r/Fuckthealtright      & 14,534                                \\
r/antifeminists        & 9,257                                 \\
r/facepalm             & 378,051                               \\
r/4chan                & 39,957                                \\
r/tifu                 & 203,652                               \\ 
r/RoastMe              & 152,008                              \\ \hline
\textbf{Total}         & \textbf{1,103,500}                    \\ \hline
\end{tabular}}
\caption{Dataset statistic}
\label{table1}
\end{table}

\noindent \textbf{Ethical Consideration} Our study does not use any user information in our analysis, and the presented results are all shown at an aggregated level, which cannot be traced back to a specific user in any way. Moreover, all the user information is anonymized by obscuring their unique IDs to protect their privacy.

\subsection*{Toxicity Analysis} \label{sec:toxicity_analysis} 

We compute the toxicity of each node and then determine the toxic accumulation for each node. To evaluate the toxicity of the responses, we utilized an off-the-shelf RoBERTa \cite{liu2019roberta} trained on the ToxiGen dataset\footnote{Publicly available at \url{https://huggingface.co/tomh/toxigen_roberta}}\cite{hartvigsen-etal-2022-toxigen}. ToxiGen is a large, well-balanced, adversarially generated corpus of implicit toxic language that can identify subtle toxic content and has the ability to differentiate between different targeted identities, thereby reducing bias. However, to ensure ToxiGen's suitability for our purpose, we tested its efficacy with Google's Perspective API\footnote{\url{https://perspectiveapi.com/research/}}. We sampled 70 posts totaling 103,967 responses from the dataset and computed toxicity scores using ToxiGen and Perspective API. Subsequently, in order to assess the extent of agreement between the aforementioned models, we calculated Cohen's $\kappa$, resulting in a value of 0.6, indicative of a moderate level of agreement. On inspection of the disagreements, interestingly, we find that ToxiGen performs extremely well in implicit hate; for example, for the text \textit{"Allanus Whoreisette"}, Perspective API score was 0.02 whereas ToxiGen predicted a score of 0.98. As a result of increased effectiveness for implicit hate, every node of a post, including the root node, was assigned a "toxicity score" computed using ToxiGen and logged for further analysis. Using this score as a starting point, we calculated the proposed toxic accumulation \textit{(Section \ref{sec:conversation_modeling})} of all of the responses in our corpus. The toxicity and toxic accumulation scores drive our analysis further.

%% file: rq1.tex
\section{Relationship between toxicity and toxic accumulation}\label{sec:rq1}

In this section, we investigate \textit{“RQ1: Are toxic comments more likely to generate toxic conversations than non-toxic ones?”}. The general connotation is that a toxic comment will cause the target to withdraw from the conversation \cite{kang2016strangers}. This suggests that, in the context of a public conversation, a toxic response has the potential to alter the conversation's content and direction as users may or may not react to toxic responses. Given toxic accumulation's ability to aggregate toxicity scores across the entire sub-branch, it serves as a dependable measure for estimating the toxicity produced by a response, thus making it the preferred approach for addressing this research question.

\subsection{Methodology}

To answer RQ1, we explored correlation statistic. Firstly, we collect all responses ignoring the hierarchy from the dataset, along with its toxicity and toxic accumulation scores. Then we calculate the Pearson correlation coefficient between the two defined scores across the entire dataset. If the response is toxic, this statistical metric can be used to ascertain whether or not the sub-branch is likely to be toxic as a result of the toxic accumulation of the node. Since the toxicity and toxic accumulation will be the same in the terminal responses (leaf nodes) we excluded them to avoid confounding our results.

\subsection{Results}

The correlation between the toxicity and toxic accumulation of the comments is presented in Table \ref{table2}. The average correlation score is 0.631 $\pm$ 0.013, indicating a moderate degree of association. The correlation is statistically significant (P-Value $<$ 0.01). The top two subreddits concerning this metric (r/tifu, and r/facepalm) are in no way controversial, unlike subreddits that are strongly centered around topics such as r/atheism, r/antifeminsts, and r/Fuckthealtright denoting the presence of variety in the data. By including the leaf nodes, the correlation turns out to be 0.9 for the entire dataset, demonstrating that leaf nodes do, in fact confound the results.

\begin{table}[h]
\centering
\resizebox{.85\columnwidth}{!}{
\begin{tabular}{lc}
\hline
\textbf{Subreddit} & \textbf{Correlation} \\ \hline
r/tifu & 0.649 \\
r/facepalm & 0.643 \\
Entire Dataset & 0.641 \\
r/athesim & 0.638 \\
r/confidentlyincorrect & 0.628 \\
r/Fuckthealtright & 0.627 \\
r/antifeminists & 0.613 \\
r/4chan & 0.610 \\ \hline
\end{tabular}}
\caption{Correlation between toxicity and toxic accumulation}
\label{table2}
\end{table}

In light of these results, it becomes evident that toxic comments have a significant influence on the overall tone and trajectory of online discussions as they possess the potential to escalate and perpetuate toxicity within conversations, leading to more negativity and hostility on social media platforms.


%% file: rq2.tex
\section{Contextual Analysis}\label{sec:rq2}

One of the main benefits of preserving the structure of a public conversation is the ability to analyze the influence of previous responses on the target response effectively. This allows us to answer the question \textit{"RQ2: Does the toxicity of a response depend on the responses made before it?"} This analysis is focused solely on the toxic nature of responses and does not consider the content or meaning of previous responses i.e, we give importance to toxic context (score) over linguistic context (text). Figure \ref{fig1} shows an example of an excerpt from a post to highlight the importance of context in terms of toxicity. Note that toxic nodes yield more toxic responses in the given example.

\subsection{Methodology}

We examine whether previous responses impact the toxicity of the target response. To achieve this, we collect toxicity scores of up to 5 previous comment levels of every response in the dataset. For example, in Figure \ref{fig1}, for the response labeled 2.1, the immediate predecessor is labeled 2, and the extended predecessor is labeled 0 and is termed level 2 analysis since we use only two levels of context, i.e, we understand how the toxicity of 0 and 2 affect 2.1. Analyzing the relationship between the toxicity of the predecessor and the toxicity of the extended predecessor demonstrates its effect on a response. Similarly, levels 3, 4, and 5 follow the same pattern, but the context becomes more in-depth. We modeled four simple linear regression models: the target variable is a response's toxicity, and the dependent variables are the responses' preceding comments.

\begin{equation}
\centering\resizebox{.75\columnwidth}{!}{
    $T = \beta_0 + \sum_{i}^{\# Levels} \beta_i T_{Level_i}$
}
\label{eq3}
\end{equation}

Equation \ref{eq3} represents a simple linear regression model, $Level_i$ refers to the previous levels' toxicity scores. $i$ takes in the levels for which the target variable is modeled. For instance, a level $2$ analysis would take $i$ as $1$ and $2$ where $1$ is parent and $2$ is a grandparent.

\subsection{Results}

Table \ref{table3} presents the beta coefficients of the models, capturing valuable insights. It is evident that the immediate predecessor holds a greater influence on the toxicity of the response compared to previous levels, while the impact of predecessors' toxicity remains minimal. Despite the low significance, we can observe a consistent downward trend in the impact levels, suggesting that the previous response plays a role in shaping the toxicity of a response within a public conversation. Similar results were obtained when analyzing individual subreddits utilizing the same methodology. Due to strong overlapping similarity, we don't report the subreddit-wise beta coefficients in the paper.

\begin{table}[htbp]
\resizebox{.95\columnwidth}{!}{
\begin{tabular}{llllll}
\hline
\textbf{Levels} & \textbf{n-1} & \textbf{n-2} & \textbf{n-3} & \textbf{n-4} & \textbf{n-5} \\ \hline
\textbf{2}      & 0.1274       & 0.0679       &              &              &              \\
\textbf{3}      & 0.1236       & 0.0898       & 0.0334       &              &              \\
\textbf{4}      & 0.1155       & 0.0925       & 0.0385       & 0.0375       &              \\
\textbf{5}      & 0.1138       & 0.0974       & 0.0316       & 0.0460       & 0.0248       \\ \hline
\end{tabular}}

\caption{Linear significance of previous levels' toxicity on comment's toxicity}
\label{table3}
\end{table}

%% file: rq3.tex
\section{Temporal Behavioral Change}\label{sec:rq3}

\begin{figure}[t]
\centering
\includegraphics[width=\linewidth]{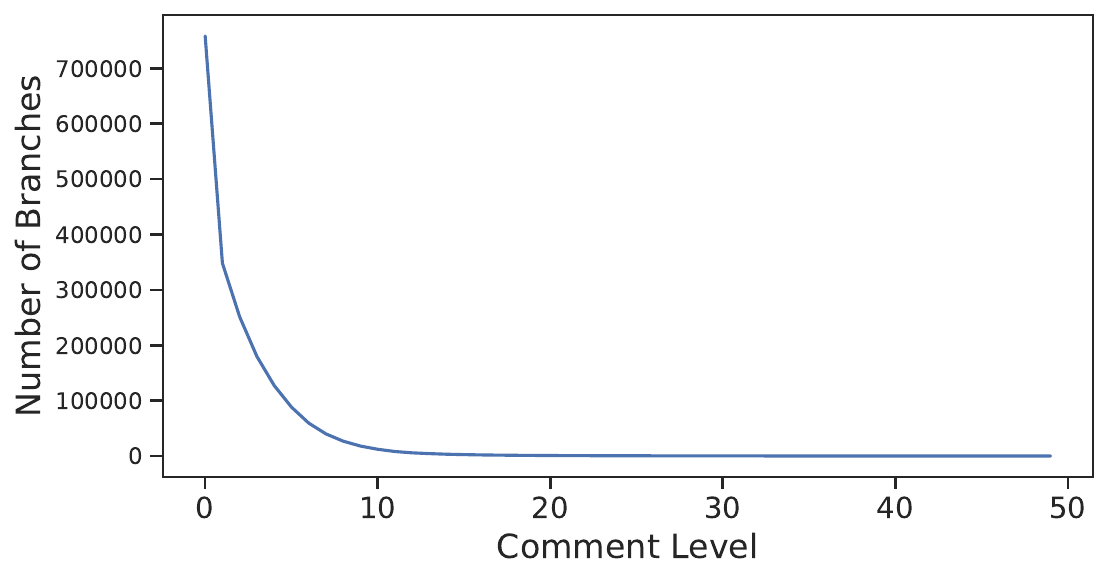}
\caption{Total number of conversational branches}
\label{fig4}
\end{figure}

This section examines the characteristics of public conversations in terms of change with respect to time. Here, time corresponds to the depth of a conversation's branches. We examine the duration of toxic prevalence in a conversation, answering the question \textit{"RQ3: How long does a conversation retain its toxic nature?"}. We find that toxic conversations are generally lesser in prevalence than non-toxic conversations.

\subsection{Methodology}

Each branch, characterized by its toxicity scores, forms a sequence of numbers where each value depends on the preceding levels. This configuration allows us to treat the data as time series, offering insights into the evolving toxicity levels as the conversation progresses. The distribution of conversational branches concerning comment levels in the dataset is illustrated in Figure \ref{fig4}. A consistent decline in the number of branches is evident beyond ten levels, with only 26,958 out of 758,519 branches extending beyond this threshold. To address RQ3, we establish a ten-level threshold for analysis. Subsequently, the mean toxicity scores and accumulated toxicity scores for selected branches are computed and visually examined to discern any trends in the rate of change concerning toxicity.

\begin{figure*}[ht]
    \centering
    \subfloat[]{\includegraphics[width=0.4925\textwidth]{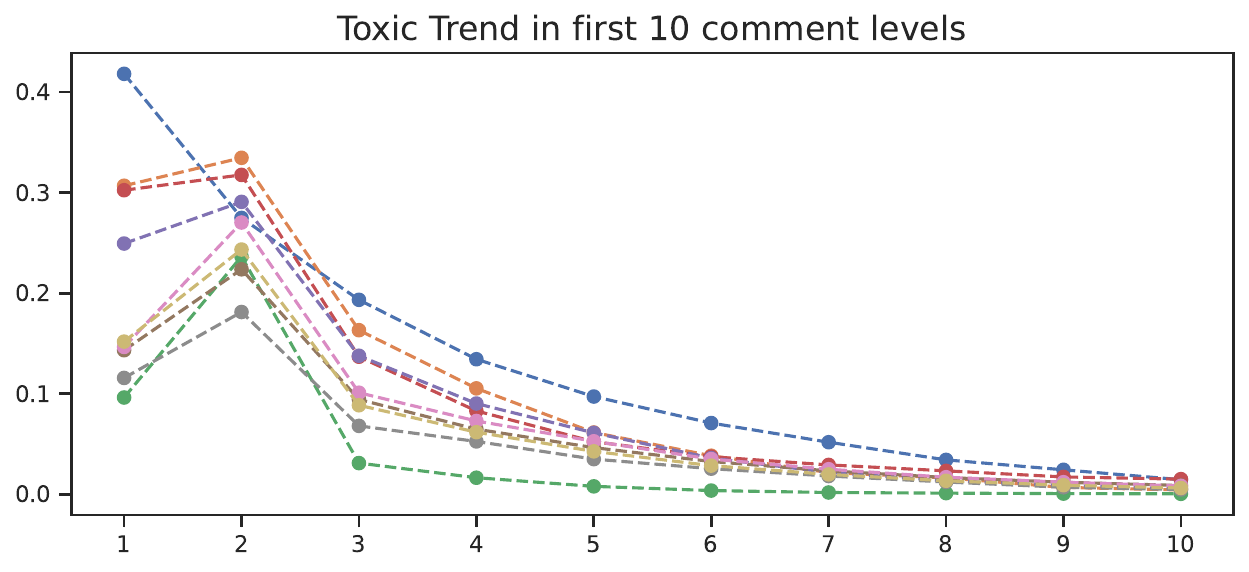}\label{fig3:A}}\hfill
    \subfloat[]{\includegraphics[width=0.4925\textwidth]{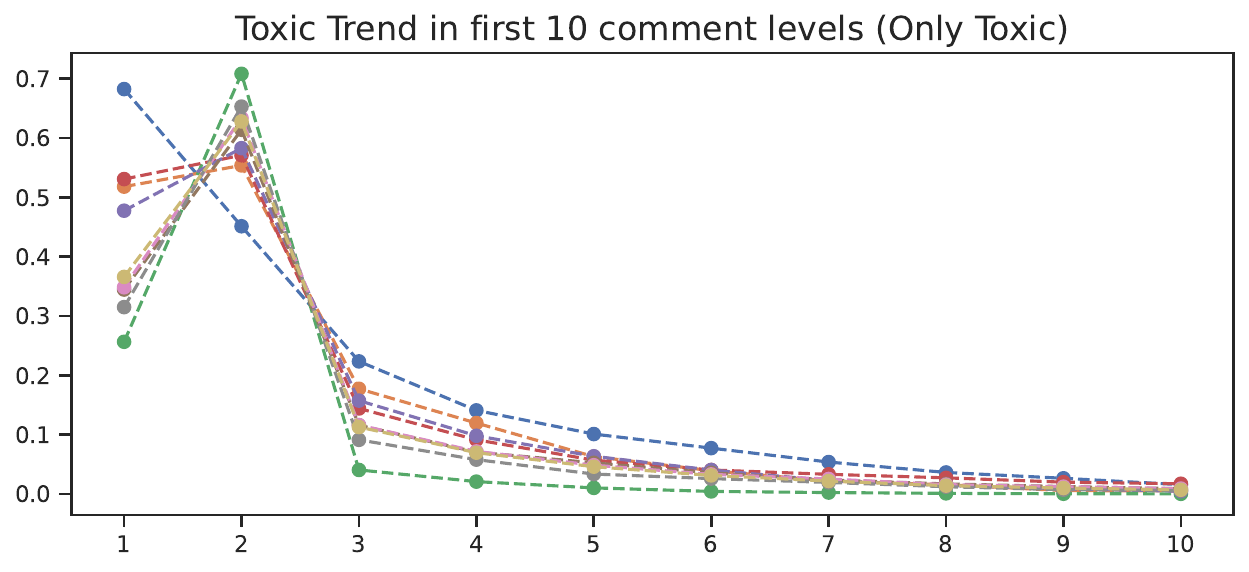}\label{fig3:B}}\\
    \subfloat[]{\includegraphics[width=0.4925\textwidth]{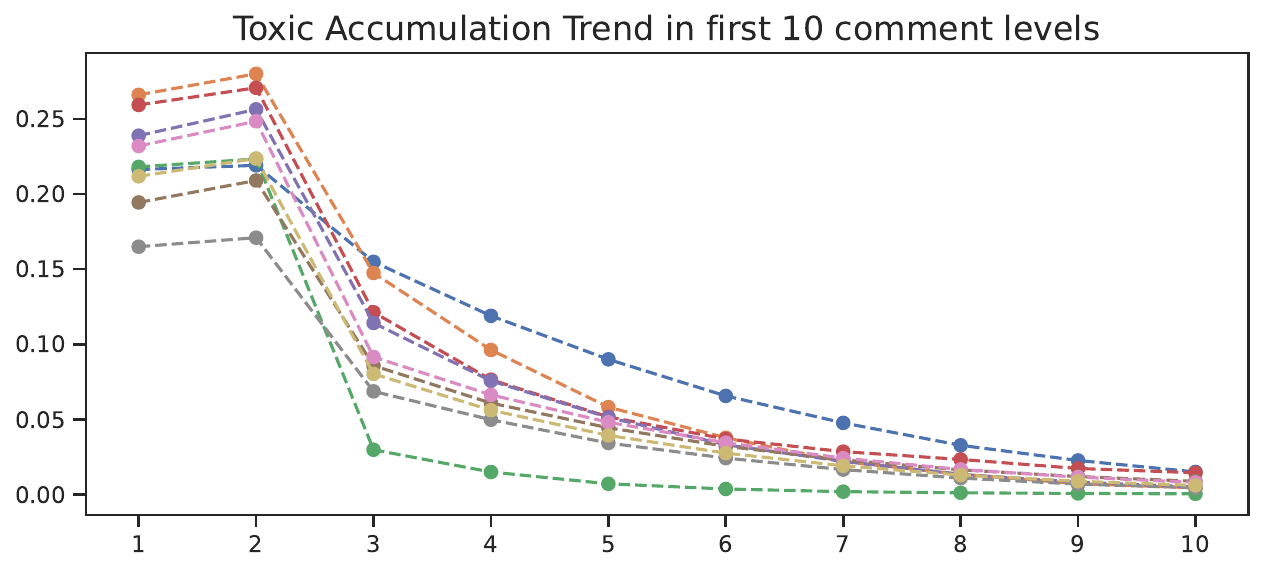}\label{fig3:C}}
    \subfloat[]{\includegraphics[width=0.4925\textwidth]{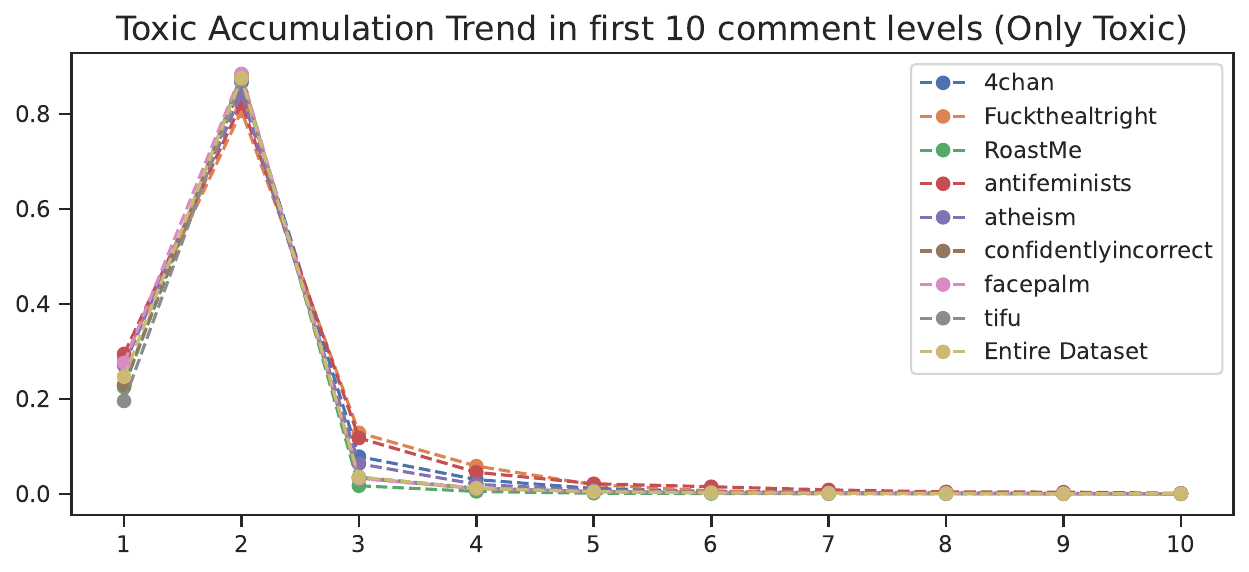}\label{fig3:D}}
    \caption{Toxic trends of conversations. X-axis denotes the comment level and Y-axis denotes the mean toxic metric}\label{fig3}
\end{figure*}

\subsection{Results}

Figure \ref{fig3} illustrates the trends of toxicity and toxic accumulation over time. Subfigures \ref{fig3:A} and \ref{fig3:C} communicate the trends irrespective of the nature of the posts, whereas, Subfigures \ref{fig3:B} and \ref{fig3:D} describe the toxic trends for posts whose toxic threshold cross 0.5. It can be observed that after the first three levels, the toxicity tends to level off irrespective of the comment being toxic or not toxic (toxicity score $>$ 0.5), signifying that conversations in public setting generally does not hold shape for long. This also resonates with the harmful effects of toxicity, as toxic comments have the effect of causing users to lose interest in the content and move on from the conversation \cite{schmid2022social}. Interestingly, the subreddit r/4chan is slightly distinct from others because, in contrast to other subreddits, the toxic levels are generally high, decrease steadily, and never fluctuate. It is likely the result of reposting content onto this subreddit from the platform 4chan, which is known for its notorious toxic levels due to its anonymity.

%% file: rq4.tex
\section{User participation with toxic comments}\label{sec:rq4}

\begin{figure*}[t]
\centering
\includegraphics[width=\linewidth]{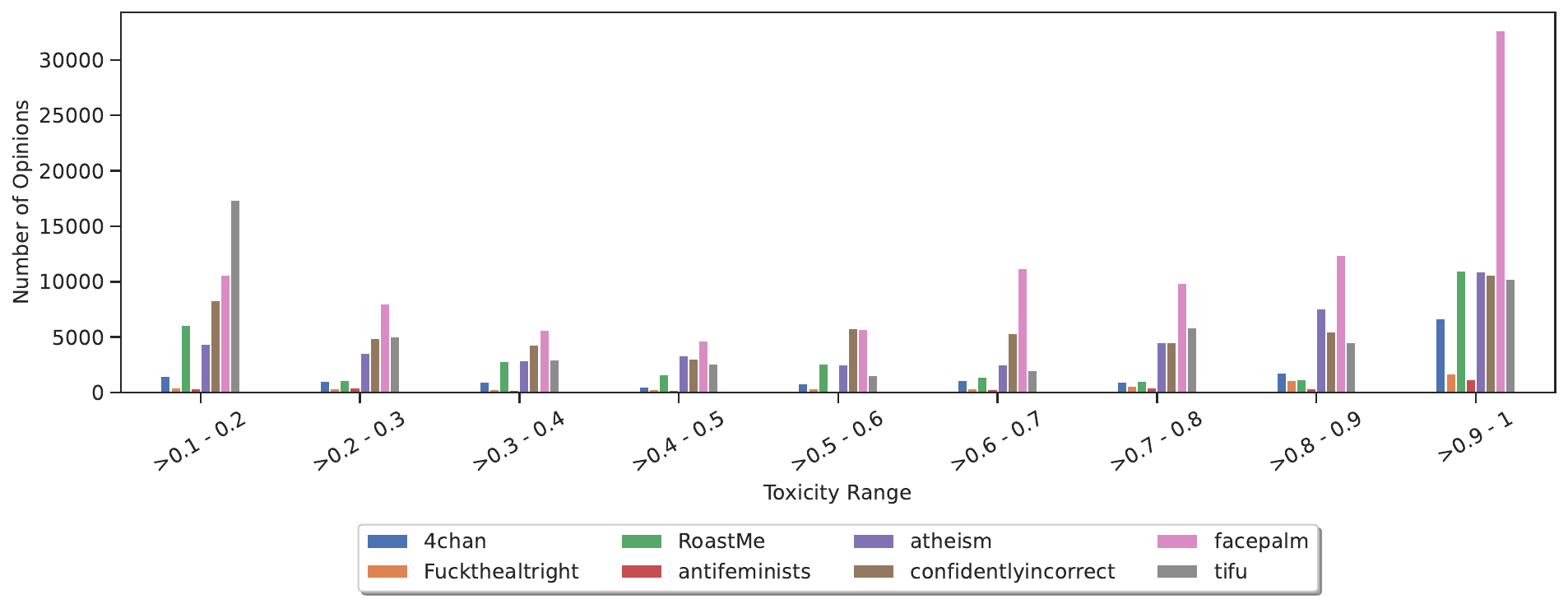}
\caption{Opinions generated on different levels of toxicity}
\label{fig5}
\end{figure*}

This section undertakes an investigation into the repercussions of toxicity on user retention in the context of public conversations. To achieve this objective, we employ the definitions of engagement and opinion as delineated in Section \ref{sec:conversation_modeling} to ascertain the influence of these variables on user participation. Through a very simple analysis of the intricate interplay among toxicity and opinions, we derive rudimentary insights pertaining to the ramifications of toxicity on user retention within the conversational realm, thus addressing the research question posited as \textit{"RQ4: What effect does toxicity have on user retention within a conversation?"}.

\subsection{Methodology}

To begin our analysis, we categorize the spectrum of toxicity into ten groups with a step range of 0.1, with group 1 representing non-toxic and group 10 representing highly toxic, enabling a more granular examination of the toxicity levels under consideration. Following this, we take the sum of the number of opinions garnered by all these ten groups in the dataset. Due to the significant disparity in the dataset regarding toxicity levels $<$ 0.1, indicating a substantial portion of non-toxic responses on social media, the outcome could be heavily biased. As a result, we made the decision to remove all data points associated with group 1.

\subsection{Results}

Figure \ref{fig5} illustrates the total number of responses categorized by toxicity range. In the plot, one can observe the presence of bimodal phenomena. This phenomenon indicates that the number of opinions increases when the text is either too toxic (toxic score $>$ 0.8) or normal. This further adds to the reasoning that people prefer to continue responding to non-toxic texts that adhere to positive and respectful communication norms rather than engaging with toxic content. However, we also observe a counter-intuitive result that the number of responses increases as the toxicity increases. Upon further inspection of parent-child response pairs, we find that out of 190,929 toxic parent responses 27.9\% children responses are toxic whereas out of 911,771 non-toxic parent responses only 19.3\% children responses are toxic signifying that toxic responses has better chance of inviting more toxic responses.

%% file: rq5.tex
\section{Effect of Consensual Profanity} \label{sec:cp}

Abusive language is not just used to attack individuals; it may also be used to frame profane information in civil discourse \cite{jordan2020profanity}. We analyze consensual toxic comments and compare them with unsolicited toxic and hateful content which are not consensual to answer \textit{"RQ5: Does the toxicity norms of consensual groups differ from non-consensual groups?"}. To this end, we scrape the responses of the top 100 posts from r/RoastMe, totaling 152,008 responses from the temporal distribution as our dataset. We chose r/RoastMe since it is a community where users post photos of themselves asking to be \textit{roasted} (severe criticism) by other members of the community. We performed the same set of analyses mentioned in the previous sections to understand if consensual toxicity patterns resemble the general toxic norms in public discussions.

The correlation between toxicity and toxic accumulation for the responses belonging to this subreddit is 0.66, which is slightly higher than that of the non-consensual subreddits (Refer to Table \ref{table2}). However, the difference is a minuscule +0.03 from the mean correlation. The contextual analysis on r/RoastMe resulted in the same significance pattern of the beta coefficients of the predecessors signaling that the immediate predecessor's toxicity impacts a node more than extended predecessors for the non-consensual communities as well. The mean scores of toxicity and accumulated toxicity resulted in the same trend overlapping with the patterns observed in Figure \ref{fig5}. The total number of opinions generated 141,757 responses for toxicity levels $<$ 0.3 \& $>$ 0.8 and 10,251 for toxicity in the range [0.3, 0.8]. The results help us in establishing the similarity between the toxic norms of consensual and non-consensual communities.

%% file: conclusion.tex
\section{Conclusion} \label{sec:conclusion}

In this paper, we comprehensively analyze toxicity in public conversations on online social media platforms. We examined 800 Reddit posts with over 1 million responses using a tree hierarchy structure. We determined the toxicity of each response using a RoBERTa model that was pre-trained on an adversarially generated dataset specifically designed to combat subtle toxicity. Our findings suggest that toxic comments have multiple negative effects on the continuation of a public conversation. For instance, we observed that the presence of toxicity in a conversation increases the likelihood of future toxic comments and discourages users from continuing to participate. An observation from our study is that users tend to interact less with responses when the content is neutral when compared to toxic or non-toxic content. Interestingly, consensual and non-consensual toxic communities follow the same toxic norm patterns.

\noindent \textbf{Limitations} In order to ensure that our dataset is as unbiased as possible, we took extra care in selecting it. Despite our efforts, however, there may still be some degree of bias present in the dataset. 

\section*{Ethical Consideration}
Our research does not use any user information in our analysis, and the presented results are all shown at an aggregated level, which cannot be traced back to a specific user in any way. Moreover, all the user information is anonymized by removing or obscuring their unique IDs to protect their privacy. 
